\documentclass{cta-author}

{}
{}
{}
\usepackage{subfig}
\usepackage{framed}

\begin{document}
\supertitle{Short Communication}

\title{Monitoring of people entering and exiting
	private areas using Computer Vision}

\author{\au{Vinay Kumar V $^{1}$}
	and 
\au{P Nagabhushan $^2$}
}

\address{\add{1}{Department of studies in Computer Science, University of Mysore, Mysuru, India}
\add{2}{Indian Institute of Information Technology - Allahabad, Prayagraj, UP, India}
\email{vkumar.vinay@ymail.com}}

\begin{abstract}
Entry-Exit surveillance is a novel research problem that addresses security concerns when people attain absolute privacy in camera forbidden areas such as toilets and changing rooms that are basic amenities to the humans in public places such as Shopping malls, Airports, Bus and Rail stations. The objective is, if not inside these camera forbidden areas, from outside, the individuals are to be monitored to analyze the time spent by them inside and also the suspecting transformations in their appearances if any. In this paper, firstly, a pseudo-annotated dataset of a laboratory observation of people entering and exiting the camera forbidden area captured using two cameras in contrast to the state-of-the-art single-camera based EnEx dataset is presented. Conventionally the proposed dataset is named \textbf{\textit{EnEx2}}. Next, a spatial transition based event detection to determine the entry or exit of individuals is presented with standard results by evaluating the proposed model using the proposed dataset and the publicly available standard video surveillance datasets that are hypothesized to Entry-Exit surveillance scenarios. The proposed dataset is expected to enkindle active research in Entry-Exit Surveillance domain.
\end{abstract}

\maketitle

\section{Introduction}
Intelligent Video surveillance systems \cite{1} contribute to realization of the concept of smart cities by addressing the public safety issues. They provide scope to track every person and analyze every event in their ambit. With the growing availability of economic surveillance cameras and processors, wide area surveillance using multiple cameras bring entire area in public places such as malls, airports, bus and rail stations under surveillance to enhance public safety. However, there are certain places like wash rooms, baby feeding rooms and changing rooms, referred to as private areas where camera surveillance is completely forbidden to ensure absolute privacy to the public. This is considered as a major setback to security systems. Suspicious events such as crimes and medical emergencies are left unnoticed for a longer time. Hence, there is a need for surveillance system that can facilitate monitoring of persons entering into / exiting from private areas without intruding the privacy of the public by analyzing their time of entry and exit, transformation in their appearances between before and after attaining absolute privacy.
The notion of Entry-Exit Surveillance (EES) provides scope for monitoring people entering and exiting private areas by placing the cameras outside. It is essentially a tracking problem where every person who enter private area is tracked for a convinced exit. This involves various individual sub problems as discussed below.
\begin{itemize}
	\item People detection - Detection of every person who enter the field of view.
	\item Motion based people tracking throughout their appearance in the field of view to confirm if they enter the private area or come out of the private area.
	\item Learning and representation of features of every person who enter the private area to re-identify when they exit.
\end{itemize}

\subsection{Progress in Entry-Exit Surveillance and Challenges}
The problem has received relatively little attention. Traces of research available in the literature include \cite{2,3,4}.  
\begin{itemize}
	\item \cite{2} introduces a benchmark dataset- EnEx dataset that demonstrates various dimensions of the problem. The dataset comprise of video sequences captured using single camera in five different locations by placing the camera so as to view the entrances of the private areas. The camera could capture back view of the individuals during entry and front view during exit. This flipped view difference of an individual between entry and exit events made re-identification challenging. State of the art Person re-identification algorithms suffered even without variations in appearances of the individuals due to the asymmetry between the flipped view of the individuals. 
	\item The method in \cite{3} analyze the direction of propagation of distortion in the color and intensity channels of the video signal to detect entry and exit events. However, the method	is limited to ideal situations like one individual entering or exiting at an instance which is not always the case in real-time applicability scenarios.
	\item The model proposed in \cite{4} focus on endorsing exit of every person who had entered private area by re-identifying them using appearance cues. 
\end{itemize}
\subsection{Motivations and Contributions}
From the above discussion, it can be concluded that
the EnEx dataset proposed in the literature is considered as a kick start for addressing Entry-Exit surveillance problem but is limited due to single view that in turn limits the performance of the state of the art algorithms due to the real-time challenges such as occlusions and wide variety of outfits of individuals. Hence, there is a need for a better performance evaluation benchmark that provide $ 360^{o} $ view of the individuals using multiple cameras to enhance re-identification accuracies. This also helps in improvising \textit{entry-exit event detection} accuracies as occlusions caused by overlapping individuals is handled using multiple camera views.
\par In this backdrop, following are the major contributions presented in this paper.
\begin{itemize}
	\item A new dataset of a laboratory observation of people entering and exiting the private areas using two cameras is proposed. The dataset is conventionally named \textit{\textbf{EnEx2}} to indicate dataset for Entry-Exit Surveillance using two cameras. 
	\item Baseline results are presented for Entry-Exit event detection method that is proposed to analyze the three dimensional layout of the camera view as well as the transition of individuals through frames and to determine the type of event as \textit{entry} or \textit{exit} or \textit{just appeared}. 
\end{itemize}

\section{The Proposed System}
This section discusses the newly created dataset in the first part and the proposed Entry-Exit event detection method in the next part.

\subsection{EnEx2 Dataset}
The dataset consist of video sequences captured in 2018 using two cameras, simulating entry and exit of people into/from the private area as well as people crossing the field of view of the camera in a continuous corridor without entering the private area. This experiment used Master degree student volunteers from a university in India.  
Two IP cameras of resolution $ 1920 X 1080 P $ are placed opposite to each other at a height of two meters from the ground among which one camera has the entrance of the private area in its field of view. The entrance on top of it has one camera placed and the opposite camera viewing the entrance is placed at a distance of 9.5 meters with oblique view. The camera cannot be placed parallel to the entrance to have the view of the entrance and placing it orthogonal to the entrance makes the public privacy compromised. Hence, oblique view of the entrance is chosen. 
The different camera views that make up the dataset are
shown in Figure . The work presented in this paper is based on the view in the figure \ref{fig:1a}. 

\begin{figure}[htbp]
	\centering
	\subfloat[Entrance view.\label{fig:1a}]{\includegraphics[width=0.3\textwidth]{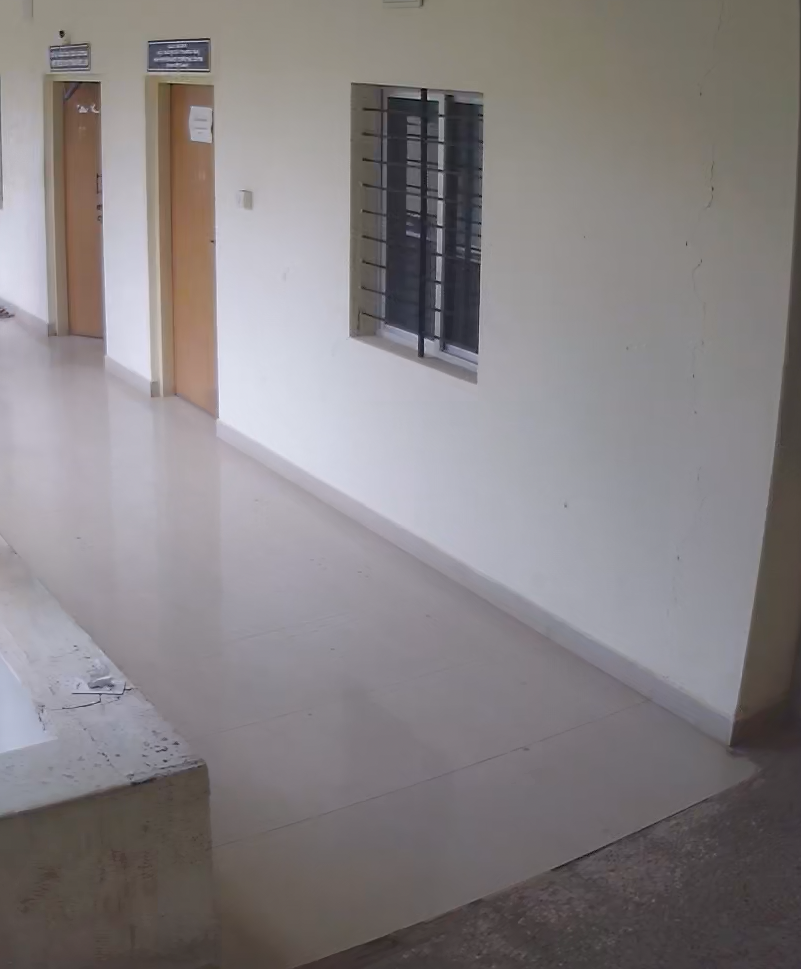}}\hfill
	\subfloat[Opposite view.\label{fig:1b}] {\includegraphics[width=0.3\textwidth]{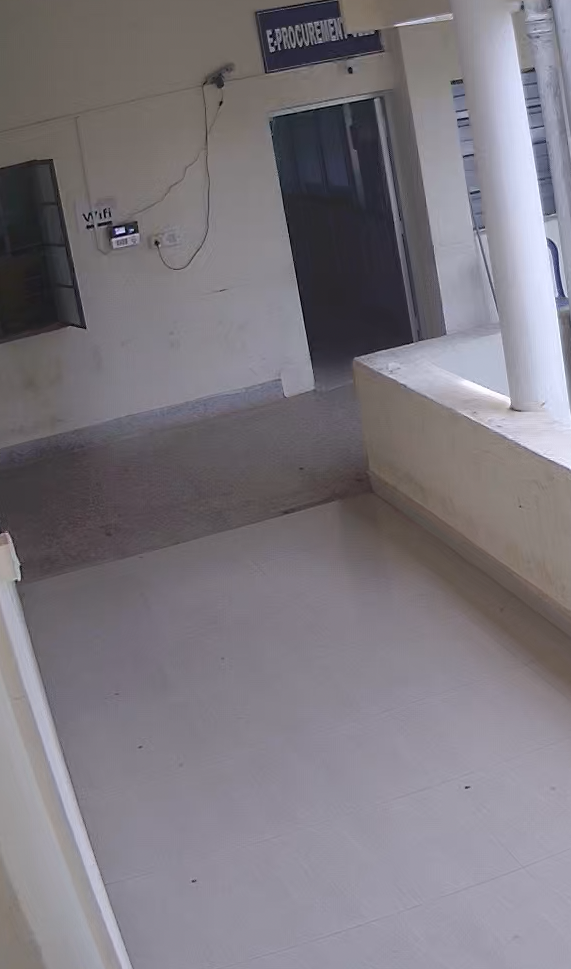}}\hfill

	\caption{Sample background images of video sequences in EnEx2 dataset}
	
	\label{fig:1}
\end{figure}
\par The working hypothesis is that every individual appearing in the field of view is to be tracked for their origin and shrink positions to study the behavior of individuals in the entry-exit surveillance ambit and its relation with the private area to detect the entry and exit events. Every individual who is detected to have entered the private area is represented with a set of features in the gallery and every individual who is detected to have exited the private area is
considered as probe to match with the gallery. The matched individual is to be removed from the gallery.
The video clips are of duration one minute each captured at the rate of 20 frames per second (fps). The videos are captured for four hours each on four consecutive days. 
One major concern in simulating the EES environment is that the individuals might be aware that they are observed by the cameras. Hence videos captured in the initial period, that is first day are not considered and after individuals getting used to the experimental set-up,  they behaved normally  which is evident in the videos in the dataset.
Tracks of every individual in the video sequences with their positions in each frame are provided as ground truth. With this the tasks of Entry-Exit event detection and person re-identification after attaining absolute privacy can be evaluated. Tracks are built using intelligent people detectors \cite{6} after subtracting frames from the background \cite{5}. The background are incrementally learnt and are updated to handle illumination variations. Every newly detected individual is tracked in the continuous frames using kalman filters \cite{8}.  

\section{Computer Vision analysis}

\subsection*{Background}
\par \textit{People Detection-} The preliminary task in Entry-Exit surveillance is to detect all the individuals in every frame. This is currently performed using static cameras and hence we review works on static camera based surveillance here. In many tracking applications, adaptive gaussian mixture model \cite{10} is used to model the background which proves to be computationally costly in real-time applicability scenarios as well as motion of the individuals is concentrated rather than segmentation of the individual completely. Gradient based features such as histogram of oriented gradient (HOG) in \cite{6} has marked substantial gains over intensity based features. With efficient background subtraction before applying the detector gains faster computation time. With the introduction of Aggregated Channel Features in \cite{7}, individual detection has attained higher accuracy even with low resolution video frames.

\textit{Tracking-} Spatial tracking of every individual in the surveillance ambit is one of the important
phases in automated video surveillance. In multi-camera video surveillance systems, tracking is of two types - intra-camera tracking where individuals are tracked within a camera view and inter-camera tracking where individuals are tracked across multiple camera views where the camera views can be overlapping or are adjacent or are disjoint. The main focus in this paper is to track individuals within a camera view. Detailed discussion on individual tracking methods can be found in \cite{9,13}.

\subsection{Problem Statement - } The aim is to spatially track every individual who enter the field of view of the camera till their departure. Two events are marked for each appearance of every individual $ S_{i} $ in the camera view-
\begin{itemize}
	\item[1] Entry into the scene- Here the task is to analyze the origin point of the individual and classify the event either as Entry into the scene $ E_{A} $ or Exit from the private area $ E_{X} $.
	\item[2] Exit from the scene- Here the task is to analyze the sink point of the individual and classify the event either as Entry into the private area $ E_{N} $ or Exit from the scene $ E_{A} $.
\end{itemize}  
The appearance of every individual who enter the camera view scene is to be marked either as entry or exit or just appeared events based on the following transitions.
\begin{itemize}
	\item $ ( E_{A}, E_{A}) $ - \textit{Just appeared}.
	\item $ ( E_{A}, E_{N}) $ - \textit{Entry}.
	\item $ ( E_{X}, E_{A}) $ - \textit{Exit}.
	\item $ ( E_{X}, E_{N}) $ - \textit{Re-entry (Just appeared)}.
\end{itemize}

\subsection{Event detection }
\begin{figure}[htbp]
	\centering
	\subfloat[$ E_{A} $\label{fig:EA}]{\includegraphics[width=0.21\textwidth]{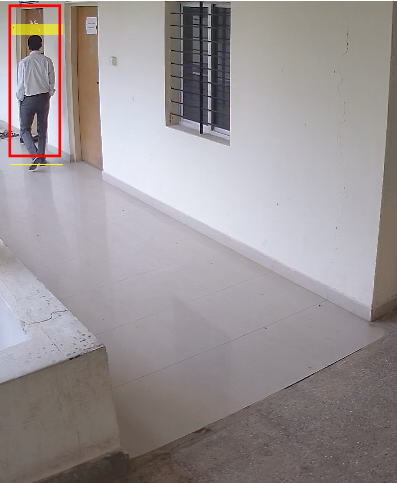}}\hfill
	\subfloat[$E_{N} $\label{fig:EN}] {\includegraphics[width=0.21\textwidth]{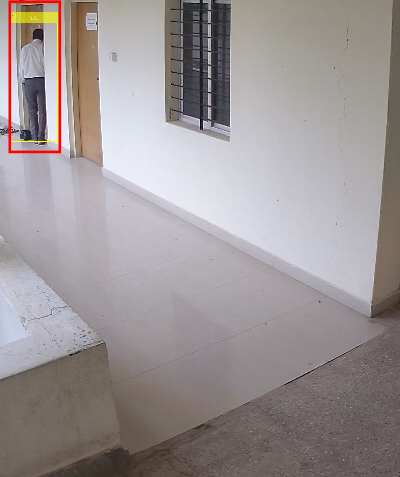}}\hfill
	
	\caption{Illustration of transition of individuals from state $ E_{A} $ to $ E_{N} $}
	
	\label{fig:2}
\end{figure}

This is illustrated with an example of an entry sequence as shown in figure \ref{fig:2}.

Let the area covered by entire area be $\textit{A}$ with coordinates $ A[0,0,x_{A},y_{A}] $ where $x_{A}$ and $y_{A}$ are width and height of the frame respectively. Let the entrance of the private area be represented by $\textit{E}$ with coordinates E$[x_{Emin},y_{Emin},x_{E},y_{E}]$ where $x_{Emin}$ and $y_{Emin}$ are top-left coordinates of the entrance in the frame and $x_{E}$ and $y_{E}$ are width and height of the entrance respectively. The entrance is marked in \textit{red} for EnEx2 dataset as shown in the figure \ref{fig:2}. 

\par Let there be $\textit{n}$ individuals who appears in the camera view scene in the interval $[t_{Start},t_{End}]$ be represented by $S = \{S_{1},S_{2},...,S_{n}\}$. 

\par For each individual, before creating a track, it is important to determine whether the individual entered the scene from outside or exited from private area. If the individual detected as a new track is bounded in a bounding box B (marked in \textit{yellow} in figure \ref{fig:2}) and if $ B\subset E$, then it can be concluded that the individual has just entered the scene by exiting from private area. However, if $ B\not\subset E$, then it can be concluded that the individual has just entered the scene from outside. 
\par Similarly, before marking each individual to have exited the scene,  it is important to determine whether the individual exited the scene to outside or entered private area. If the track is unassigned for a certain amount of frames, then the last frame where the individual was assigned has to be examined for its spacial location in the frame. If the individual is bounded in a bounding box B  and if $ B\subset E$, then it can be concluded that the individual has just entered the private area. However, if $ B\not\subset E$, then it can be concluded that the individual has just exited the scene to outside. 

\subsubsection*{Challenges and Solutions: } The important factors that affects Entry/Exit event detection are- overlapping of individuals while entering and exiting, and individuals occluding the entrance that makes it challenging to determine the entry/exit of other individuals into/from the private area. The solutions for these challenges are as follows.

When there are more than one individual in the scene and if one individual is blocking the entrance, then the entrance region has to be updated for that frame for other individuals. If Si is the individual who blocks the entrance and if Bi is its bounding box, then for every other individual in the scene, the regions E-B and Bi-B' can act as entrance where Bi' is the extended area of Bi. 

The multiple view provided by two cameras in the proposed dataset help in locating the individuals in a track that are occluded by individuals overlapping in the entrance facing view using the opposite view. The spatial co-ordinates of the two cameras are mapped using mapping function as part of camera calibration. 

\par The time ti and tj has to be marked for each individual as time of entry into the scene and time of exit from the scene respectively and the activity detected has to be updated as \textit{Entry} or \textit{Exit} or \textit{Just appeared} accordingly. 
\par
\begin{figure}\centering
	\begin{framed}
		\subfloat[New Track\label{fig:1a}]{\includegraphics[width=0.50\textwidth]{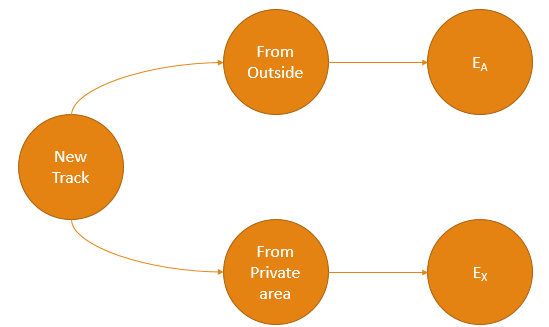}}\hfill
		\end{framed}
	\begin{framed}
		\subfloat[Missing Track\label{fig:1a}]{\includegraphics[width=0.50\textwidth]{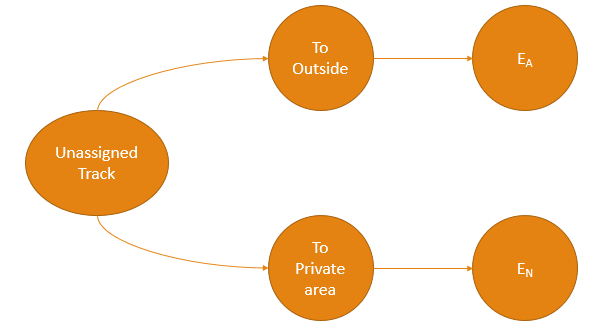}}\hfill
		\end{framed}
	\caption{State of each individual based on their position in the frames} \label{fig:3}
\end{figure}

\begin{figure}
	\centering
	\begin{framed}
	\includegraphics[width=6cm,height=5cm,keepaspectratio]{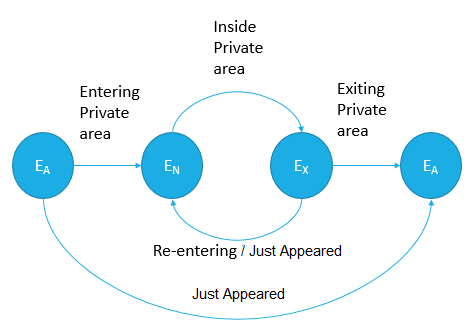}\end{framed}
	\caption{State transition diagram for tracking of individuals in Entry-Exit surveillance} \label{fig:4}

\end{figure}

 The state transitions of the individuals who appears in the camera view scene in Entry/Exit Surveillance is as shown in figures \ref{fig:3} and \ref{fig:4}.

\section{Evaluation}
The proposed Entry-Exit event detection model is evaluated using the proposed dataset as well as the standard EnEx dataset which is captured using single camera that has entrance of the private area in its field of view. In addition, the CAVIAR dataset \cite{11} that comprise of video sequences captured in a shopping mall corridor where people move in and out of a shopping outlet and the PAMELA-UANDES dataset \cite{12} where people enter and exit out of metro trains are hypothesized to Entry-Exit scenarios and the proposed method is evaluated. As the problem is novel to compare with the current state of the art, the performance of the model is evaluated across the mentioned datasets.

Given the video stream as input to the system, individuals are detected in every frame using Histogram of Oriented Gradient based deeply learned people detection method \cite{6}. For every new individual in the field of view, a new track is created and is assigned throughout its appearance in the field of view and every unassigned track after a thresholded number of frames is labeled as missing track and the frame in which the individual last appeared is considered last frame of the track. 

Each track is given as input to the proposed model. The initial and final frames of the track are analyzed and are labeled as $ E_{N} $, $ E_{X} $ and$  E_{A} $ based on their position in the frames as discussed in the previous section. The initial and final state of the individuals are represented as ordered pair and the transition of state of the individuals during their presence in the field of view determines the type of event of the track as shown in figure \ref{fig:4}.

\par The performance evaluation analyses of the proposed method that was evaluated on the above mentioned datasets are as follows. 

\begin{table}\centering
	\caption{People detection results.}\label{tab1}
	\begin{tabular}{|c|c|c|c|}
		\hline
		Dataset &  Recall & Precision & F1\\
		\hline
		EnEx \protect{\cite{2}} & 60.4  & 80.3 & 68.94\\
		\textbf{\textit{EnEx2}}  & 80.43  & 92.23 & 85.92\\
		CAVIAR \protect{\cite{11}} & 78.2  & 91.3 & 84.24\\
		PAMELA-UANDES \protect{\cite{12}} & 80.76 & 90.05 & 85.09\\
		\hline
	\end{tabular}
\end{table}

\begin{table}\centering
	\caption{Entry/Exit event detection results.}\label{tab2}
	\begin{tabular}{|c|c|c|c|}
		\hline
		Dataset &  Entry & Exit & Just appeared\\
		\hline
		EnEx \protect{\cite{2}} & 94.4  & 96.4 & 98.3\\
		\textbf{\textit{EnEx2}} & 96.2  & 98.1 & 98.7\\
		CAVIAR \protect{\cite{11}} & 93.2  & 94.3 & 93.3\\
		PAMELA-UANDES \protect{\cite{12}} & 70.6 & 62.5 & 95.09\\
		
		\hline
	\end{tabular}
\end{table}

\par People detection on all the four datasets are found to be satisfactory.The lower people detection rate in EnEx dataset as compared to the other two datasets is because of the camera placements where complete body of the individuals is not available in the frames. The missing rate in caviar dataset is due to frequent occlusion of individuals overlapping each other. However, the method performs best on CAVIAR dataset among the four datasets in classifying the events as Entry/Exit/Miscellaneous. Low Entry-Exit event detection in PAMELA- UANDES dataset is due to the top view of the scene where the visibility of the walking path of the individuals is limited. Overall, the method proposed in this paper gives satisfactory results in detecting the events in Entry-Exit surveillance.

\section{Conclusion}
In this paper, a pseudo-annotated dataset \textit{EnEx2} that consist of video sequences simulating Entry-Exit Surveillance scenarios using two oppositely placed cameras is presented. In addition, a novel Entry-Exit event detection method which is a fundamental step in Entry-Exit surveillance is proposed. Baseline results obtained by evaluating the proposed method on the proposed dataset as well as other publicly available dataset is presented. High Entry-Exit event detection accuracy on EnEx2 dataset demonstrates the importance of a second camera for Entry-Exit Surveillance.  


%


\end{document}